\pdfoutput=1

\documentclass[11pt]{article}

\usepackage[final]{acl}

\usepackage{times}
\usepackage{latexsym}

\usepackage[T1]{fontenc}

\usepackage[utf8]{inputenc}

\usepackage{microtype}

\usepackage{inconsolata}

\usepackage{graphicx}

\usepackage{svg}
\usepackage{textcomp}  
\usepackage{scalerel}  
\usepackage{lipsum}
\usepackage{xcolor}
\usepackage{tikz}
\usetikzlibrary{fit,arrows,quotes,calc,automata,positioning,backgrounds,arrows.meta,bending,shapes,snakes,matrix}
\usepackage{subcaption}
\usepackage{amsmath}
\usepackage{amsfonts}
\usepackage{amssymb}
\usepackage[OT2, T1]{fontenc}
\usepackage[greek,russian,english]{babel}
\usepackage{mathtools}
\usepackage{hyperref}
\usepackage{mdframed}
\usepackage{siunitx}
\usepackage{booktabs}
\usepackage{listingsutf8}
\usepackage{amssymb}
\usepackage{pifont}
\newcommand{\cmark}{\textcolor{green}{\ding{51}}}%
\newcommand{\xmark}{\textcolor{red}{\ding{55}}}%

\definecolor{myRED}{HTML}{FF0000}
\definecolor{myBLUE}{HTML}{00FBFF}
\definecolor{myGREEN}{HTML}{00FF00}
\definecolor{myMAGENTA}{HTML}{F600FF}

\title{Ask a Local: Detecting Hallucinations With Specialized Model Divergence}

\author{
\textbf{Aldan Creo\textsuperscript{\textasteriskcentered\textdagger}},
\textbf{Héctor Cerezo-Costas\textsuperscript{\textdaggerdbl}},
\textbf{Pedro Alonso-Doval\textsuperscript{\textdaggerdbl}},
\textbf{Maximiliano Hormazábal-Lagos\textsuperscript{\textdaggerdbl}}
\\
\\
\textsuperscript{\textasteriskcentered}Independent Author, Dublin, IE \\
\textsuperscript{\textdaggerdbl}Fundación Centro Tecnolóxico de Telecomunicacións de Galicia (GRADIANT), Vigo, ES \\
\small{
\textsuperscript{\textdagger}To whom correspondence should be addressed: \href{mailto:research@acmc.fyi}{research@acmc.fyi}
}
}

\begin{document}
\maketitle
\begin{abstract}
  Hallucinations in large language models (LLMs) --- instances where models generate plausible but factually incorrect information --- present a significant challenge for AI.

  We introduce ``Ask a Local'', a novel hallucination detection method exploiting the intuition that specialized models exhibit greater surprise when encountering domain-specific inaccuracies.
  Our approach computes divergence between perplexity distributions of language-specialized models to identify potentially hallucinated spans.
  Our method is particularly well-suited for a multilingual context, as it naturally scales to multiple languages without the need for adaptation, relying on external data sources, or performing training.
  Moreover, we select computationally efficient models, providing a scalable solution that can be applied to a wide range of languages and domains.

  Our results on a human-annotated question-answer dataset spanning 14 languages demonstrate consistent performance across languages, with Intersection-over-Union (IoU) scores around 0.3 and comparable Spearman correlation values.
  Our model shows particularly strong performance on Italian and Catalan, with IoU scores of 0.42 and 0.38, respectively, while maintaining cross-lingual effectiveness without language-specific adaptations.
  We release our code and architecture to facilitate further research in multilingual hallucination detection.
\end{abstract}

\section{Introduction}
\label{sec:intro}

The phenomenon of hallucinations in LLMs poses significant challenges, especially in multilingual settings where the complexity of language nuances can exacerbate the issue.
Detecting hallucinations in LLMs is crucial for ensuring the reliability of their outputs, as hallucinations can lead to misinformation and inaccuracies in the generated text.
In this paper, we propose a novel approach to detecting hallucinations in LLMs by leveraging specialized model divergence.

For our evaluation, we utilize a dataset of hallucinations \cite{vazquez-etal-2025-mu-shroom} tagged by human annotators across 14 different languages, including Arabic, Basque, Catalan, Chinese (Mandarin), Czech, English, Farsi, Finnish, French, German, Hindi, Italian, Spanish, and Swedish.
The examples in the dataset were generated by 38 different LLMs, with a total of 200 examples per language.

Our approach is based on the hypothesis that specialized models, each trained on a specific language or domain, can provide valuable insights into the reliability of LLM outputs.
By comparing the perplexities, and particularly the divergence between the perplexities, of a set of specialized models, we aim to identify hallucinated spans in the text.
We evaluate our method on a diverse set of languages and demonstrate its effectiveness in detecting hallucinations without any external data sources or training data.
While established approaches have been extensively explored in the literature, our novel method introduces a fresh perspective with significant potential for improvement through further research. 

\begin{table*}[h]
  \centering
  \begin{tabular}{lccccc}
    \toprule
    \textbf{Method}                            & \textbf{Manual} & \textbf{RAG-like} & \textbf{Training} & \textbf{White-box} & \textbf{Ours} \\
    \midrule
    \textbf{Parallelizability}                 & \xmark          & \cmark            & \cmark            & \cmark             & \cmark        \\
    \textbf{Reproducibility}                   & \xmark          & \cmark            & \cmark            & \cmark             & \cmark        \\
    \textbf{Independence from external data}   & \xmark          & \xmark            & \xmark            & \cmark             & \cmark        \\
    \textbf{Generalization to unseen contexts} & \cmark          & \cmark            & \xmark            & \cmark             & \cmark        \\
    \textbf{Detection on obscure assertions}   & \cmark          & \cmark            & \xmark            & \xmark             & \xmark        \\
    \textbf{Scalability to multiple languages} & \xmark          & \cmark            & \xmark            & \xmark             & \cmark        \\
    \bottomrule
  \end{tabular}
  \caption{\textbf{Method comparison.} Our approach uses model divergence to detect hallucinations in multiple languages.}
  \label{tab:comparison}
\end{table*}

\subsection{Related Work}

To date, detecting hallucinations inside a given text has been primarily done in one of four ways: (1) \textbf{manual} annotation, which is time-consuming, labor-intensive and prone to human variation \cite{niu2023ragtruth}; (2) using external data sources to verify the information --- such as retrieval-augmented generation (\textbf{RAG}) \cite{rag, mishra2024fine} --- which turns the problem into a trivial task of choosing spans of text given the ground truth, but depends on the availability of external data and quality of the retrieval system; (3) \textbf{training} a model for the specific task of hallucination detection, which requires a corpus of labeled data and may not generalize well \cite{azaria2023internal}; and (4) analyzing the features of one or many \textbf{white-box} language models, such as the perplexity of the generated text or the attention maps, which primarily depends on the quality of their training data and the model architecture \cite{malinin2020uncertainty, sriramanan2025llm}.

In this paper, we present an alternative approach that is in line with the fourth method, but addresses the limitation of the dependence on a single model's quality of training data and architecture by using a set of models that are specialized for different languages and domains.
Table \ref{tab:comparison} provides a broad comparison of our method to existing approaches.

\section{Our approach}
\label{sec:approach}

\begin{figure*}[]
  \centering
  \begin{tikzpicture}[
      node distance=0.5cm,
      inner sep=0.1cm,
      font=\sffamily\small,
    ]
    \node[draw, rectangle, minimum width=5cm, minimum height=0.3cm, inner sep=0.25cm] (question) {\textit{Who was the last king of the Spanish Habsburg dynasty?}};
    \node[draw, rectangle, minimum width=5cm, minimum height=0.3cm, inner sep=0.25cm, below=of question.south east, anchor=north east] (answer) {\textit{The last king was \textcolor{myRED}{\underline{Philip}} II of Spain.}};

    \node[inner sep=0cm, rectangle, minimum width=1cm, minimum height=0.3cm, right=1.25cm of question.east, anchor=west] (chinese) {\scalerel*{\includegraphics{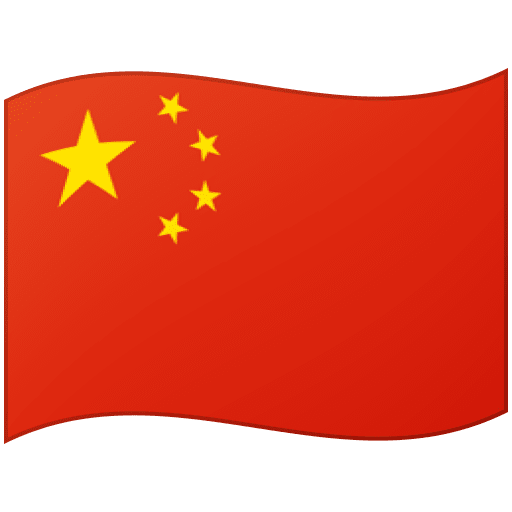}}{\textrm{\Large \textbigcircle}}};
    \node[inner sep=0cm, rectangle, minimum width=1cm, minimum height=0.3cm, below=0.1cm of chinese.south] (english) {\scalerel*{\includegraphics{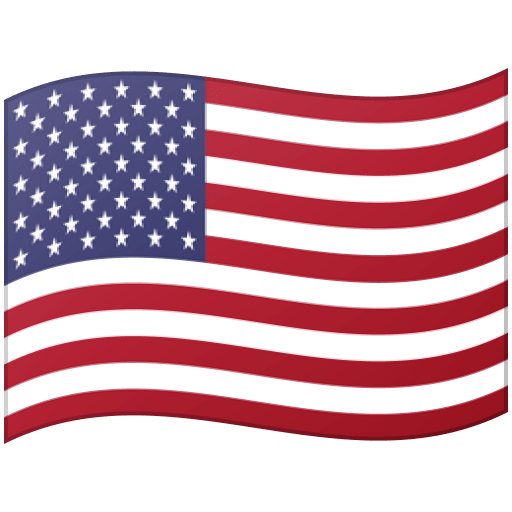}}{\textrm{\Large \textbigcircle}}};
    \node[inner sep=0cm, rectangle, minimum width=1cm, minimum height=0.3cm, below=0.1cm of english.south] (french) {\scalerel*{\includegraphics{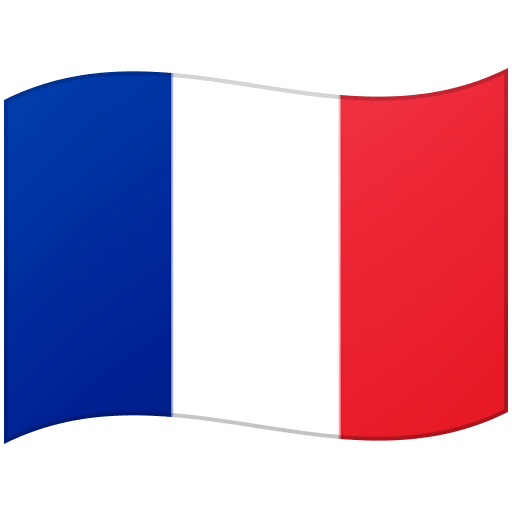}}{\textrm{\Large \textbigcircle}}};
    \node[inner sep=0cm, rectangle, minimum width=1cm, minimum height=0.3cm, below=0.1cm of french.south] (german) {\scalerel*{\includegraphics{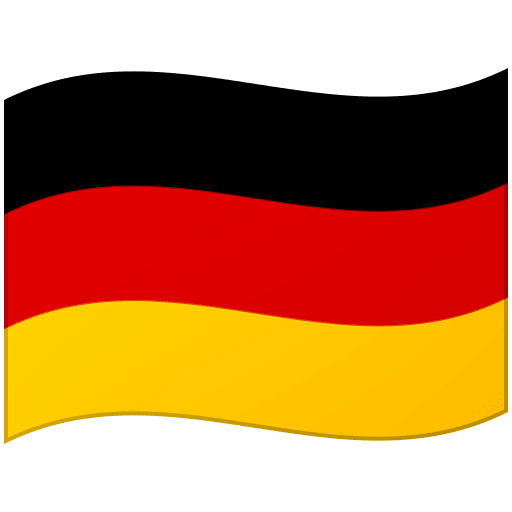}}{\textrm{\Large \textbigcircle}}};
    \node[inner sep=0cm, rectangle, minimum width=1cm, minimum height=0.3cm, below=0.1cm of german.south] (spanish) {\scalerel*{\includegraphics{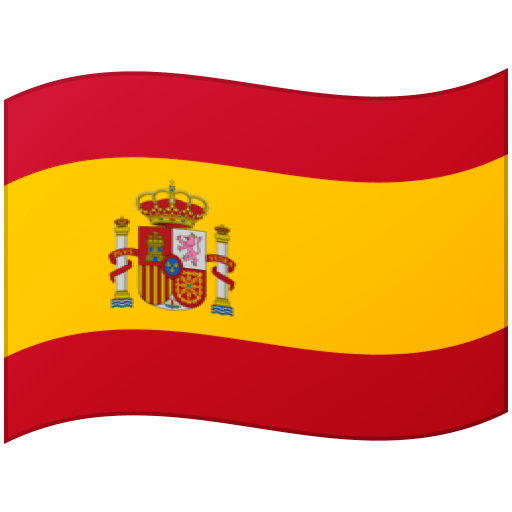}}{\textrm{\Large \textbigcircle}}};

    \node[draw, densely dashed, rectangle, fit=(french) (german) (spanish) (english) (chinese), inner sep=0.2cm] (countries) {};

    \draw[->, solid] (question.east) -- ($(question.east) + (0.3cm, 0.0cm)$) |- (french.west);
    \draw[->, solid] (question.east) -- ($(question.east) + (0.3cm, 0.0cm)$) |- (german.west);
    \draw[->, solid] (question.east) -- ($(question.east) + (0.3cm, 0.0cm)$) |- (chinese.west);
    \draw[->, solid] (question.east) -- ($(question.east) + (0.3cm, 0.0cm)$) |- (spanish.west);
    \draw[->, solid] (question.east) -- ($(question.east) + (0.3cm, 0.0cm)$) |- (english.west);
    \draw[solid] (answer.east) -- ($(answer.east) + (0.3cm, 0.0cm)$);

    \node[right=0.3cm of $(chinese.east -| countries.east)$, anchor=west, inner sep=0] (chinese_dist) {\tikz{\fill[myBLUE] (0,0) rectangle (0.02cm,0.3cm);} \vphantom{0.3cm}\raisebox{0.07cm}{\tiny{0.02}}};
    \node[right=0.3cm of $(english.east -| countries.east)$, anchor=west, inner sep=0] (english_dist) {\tikz{\fill[myBLUE] (0,0) rectangle (0.33cm,0.3cm);} \vphantom{0.3cm}\raisebox{0.07cm}{\tiny{0.33}}};
    \node[right=0.3cm of $(french.east -| countries.east)$, anchor=west, inner sep=0] (french_dist) {\tikz{\fill[myBLUE] (0,0) rectangle (0.03cm,0.3cm);} \vphantom{0.3cm}\raisebox{0.07cm}{\tiny{0.03}}};
    \node[right=0.3cm of $(german.east -| countries.east)$, anchor=west, inner sep=0] (german_dist) {\tikz{\fill[myBLUE] (0,0) rectangle (0.05cm,0.3cm);} \vphantom{0.3cm}\raisebox{0.07cm}{\tiny{0.05}}};
    \node[right=0.3cm of $(spanish.east -| countries.east)$, anchor=west, inner sep=0] (spanish_dist) {\tikz{\fill[myBLUE] (0,0) rectangle (0.58cm,0.3cm);} \vphantom{0.3cm}\raisebox{0.07cm}{\tiny{0.58}}};

    \node[draw=none, fit=(french_dist) (german_dist) (chinese_dist) (spanish_dist) (english_dist), inner sep=0.2cm] (weights) {};

    \draw[->, densely dotted] (answer.south) |- ($(countries.south) + (0.0cm, -0.2cm)$) node[near start, left] {\textit{What model should answer this?}} -| ($(weights.south) + (0.0cm, -0.0cm)$);

    \node[right=0.3cm of $(chinese.east -| weights.east)$, anchor=west, inner sep=0] (chinese_perp) {\tikz{\fill[myMAGENTA] (0,0) rectangle (0.8cm * 0.01,0.3cm);} \vphantom{0.3cm}\raisebox{0.07cm}{\tiny{0.8}}};
    \node[right=0.3cm of $(english.east -| weights.east)$, anchor=west, inner sep=0] (english_perp) {\tikz{\fill[myMAGENTA] (0,0) rectangle (0.5cm * 0.01,0.3cm);} \vphantom{0.3cm}\raisebox{0.07cm}{\tiny{0.5}}};
    \node[right=0.3cm of $(french.east -| weights.east)$, anchor=west, inner sep=0] (french_perp) {\tikz{\fill[myMAGENTA] (0,0) rectangle (6.0cm * 0.01,0.3cm);} \vphantom{0.3cm}\raisebox{0.07cm}{\tiny{6.0}}};
    \node[right=0.3cm of $(german.east -| weights.east)$, anchor=west, inner sep=0] (german_perp) {\tikz{\fill[myMAGENTA] (0,0) rectangle (20.4cm * 0.01,0.3cm);} \vphantom{0.3cm}\raisebox{0.07cm}{\tiny{20.4}}};
    \node[right=0.3cm of $(spanish.east -| weights.east)$, anchor=west, inner sep=0] (spanish_perp) {\tikz{\fill[myMAGENTA] (0,0) rectangle (69.3cm * 0.01,0.3cm);} \vphantom{0.3cm}\raisebox{0.07cm}{\tiny{69.3}}};

    \node[draw=none, fit=(chinese_perp) (english_perp) (french_perp) (german_perp) (spanish_perp), inner sep=0.2cm] (perplexities) {};

    \draw[->, densely dotted] (question.north) -- ($(question.north) + (0.0cm, 0.3cm)$) -| node[near start, above] {\textit{How perplexed is this model?}} ($(perplexities.north) + (0.0cm, 0.2cm)$) -- (perplexities.north);

    \node[right=0.2cm of spanish_perp.east, anchor=west, inner sep=0.1cm] (warning) {\tikz{\node[draw, circle, fill=red, inner sep=0.05cm] {\textbf{\textcolor{white}!}};}};

    \node[draw=myRED, rectangle, fit=(warning) (spanish_perp) (spanish_dist) (spanish), inner sep=0.0cm] (warning_box) {};

  \end{tikzpicture}
  \caption{\textbf{Our approach to detecting hallucinations.} We propose a method to detect hallucinations in a model by comparing the perplexities of a set of specialized models. The perplexity of each specialized model is computed on the same input, and the divergence between the perplexities is used to detect hallucinations.}
  \label{fig:overview}
\end{figure*}
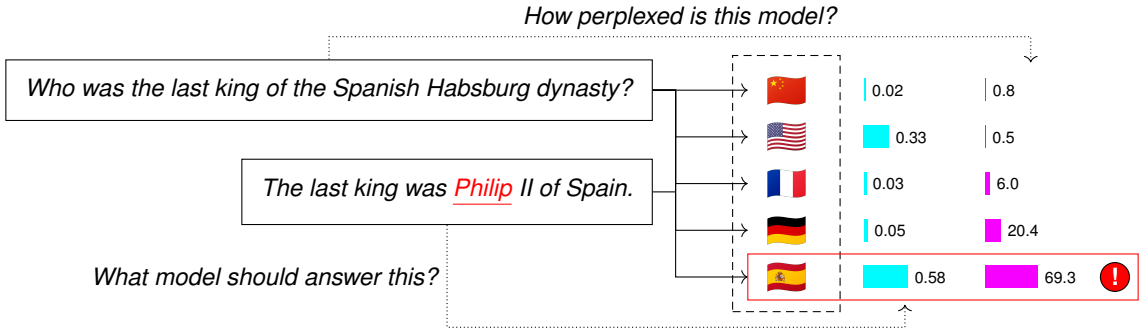

\begin{figure*}[h]
  \centering
  \begin{subfigure}[t]{0.48\textwidth}
    \centering
    \includegraphics[width=\textwidth]{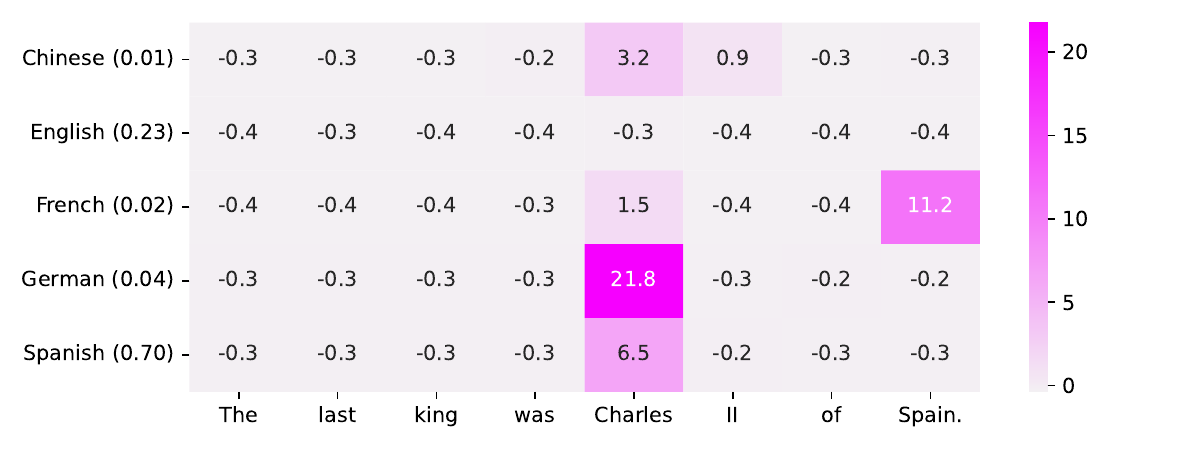}
    \caption{\textbf{Truth.} The word ``Charles'' has a higher perplexity --- which previous works associate with hallucinations. However, when we consider model divergence, the Spanish model is not significantly more perplexed.}
    \label{fig:perplexity_matrix_true}
  \end{subfigure}
  ~
  \begin{subfigure}[t]{0.48\textwidth}
    \centering
    \includegraphics[width=\textwidth]{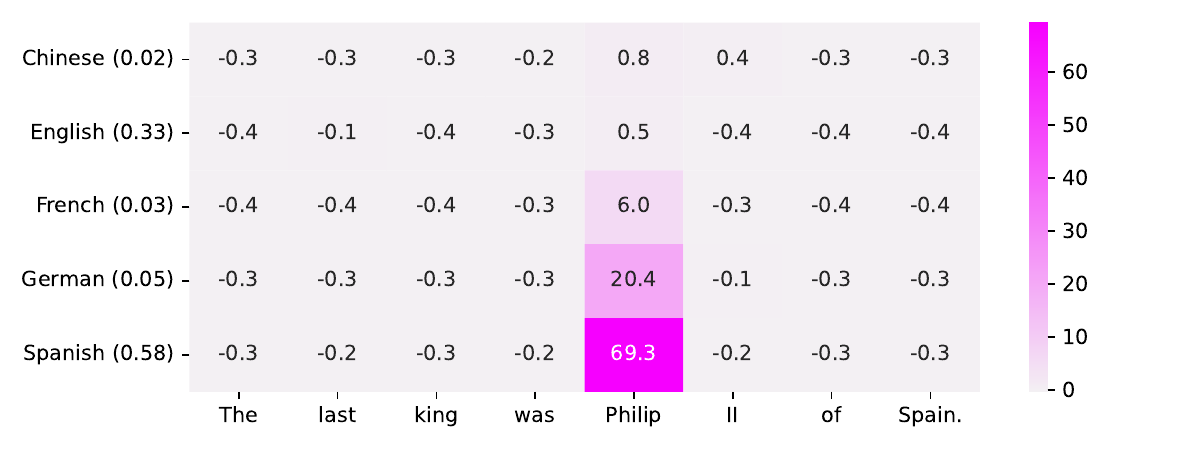}
    \caption{\textbf{Hallucination.} The model specialized in Spanish shows a significantly higher perplexity for the hallucinated word ``Philip''.}
    \label{fig:perplexity_matrix_fake}
  \end{subfigure}
  \caption{\textbf{Perplexity divergence.} We show the normalized perplexity values of the words in the answer of Figure~\ref{fig:overview} for each model in the example. We include the assigned model weights in parentheses next to the model names.}
  \label{fig:perplexity_matrix}
\end{figure*}

\begin{mdframed}[backgroundcolor=myBLUE!10]
  \paragraph{Intuition.}
  Imagine you are in a room with people from different countries.
  Someone says \textit{``The last king of the Habsburg dynasty was Philip II of Spain''}.
  Most people in the room would find it plausible --- but the Spanish person, who has some knowledge about Spain, would be more surprised: \textit{``Didn't we have a king called Charles II?''}.

  This is what we measure: is a given statement more surprising to a specialized model?

  \emph{(we formalize this intuition next)}
\end{mdframed}
Figure~\ref{fig:overview} shows a simplified illustration of our approach, while Figure~\ref{fig:perplexity_matrix} shows the perplexity matrices for that example.

\subsection{Hallucination Score}
\label{sec:hallucination_score}

We propose the following hallucination score $H$ to detect hallucinations in the text:

\begin{equation}
  \label{eq:hallucination_score}
  \begin{split}
    H(w_i) & = \beta \cdot \underbrace{\text{KL} \left( \text{PPL}_{\text{local}}(w_i) \middle\| \text{PPL}_{\text{foreign}}(w_i) \right)}_{\text{divergence in perplexities}} \\
           & + (1 - \beta) \cdot \underbrace{\sum_j^{|\mathcal{M}|} \text{PPL}(w_i; m_j)}_{\text{average perplexity}}
  \end{split}
\end{equation}

where $\text{KL}$ is the Kullback-Leibler divergence, and $\text{PPL}(w_i; {m}_j)$ is the normalized perplexity for a word $w_i$ with respect to a model ${m}_j \in \mathcal{M}$ (Sections~\ref{sec:tokenization} and \ref{sec:perplexity_scales}).
We balance the divergence and the average perplexity with $\beta$, as there may be easily detectable hallucinations where multiple models will show high perplexity.

\begin{equation}
  \label{eq:ppl_local}
  \text{PPL}_{\text{local}}(w_i) = \sum_{j}^{|\mathcal{M}|} \alpha_j \cdot \text{PPL}(w_i; {m}_j)
\end{equation}
\begin{equation}
  \label{eq:ppl_foreign}
  \text{PPL}_{\text{foreign}}(w_i) = \sum_{j}^{|\mathcal{M}|} \frac{1 - \alpha_j}{|\mathcal{M}| - 1} \cdot \text{PPL}(w_i; {m}_j)\sum_{i}^{n}
\end{equation}

Equations \ref{eq:ppl_local} and \ref{eq:ppl_foreign} compute what we call the \textbf{local} and \textbf{foreign} perplexities, respectively. $\alpha_j$ is a weight assigned to ${m}_j$, so that $\sum_{j}^{|\mathcal{M}|} \alpha_j = 1$, measuring the relevance of each specialized model to the question-answer pair (Section~\ref{sec:weights}).
Intuitively, the local perplexity is the amount of per-word surprise that we would expect from a model that is specialized in the language or domain of the text --- \emph{e.g.}, the Spanish model on Figure~\ref{fig:overview} ---, while the foreign perplexity is the amount of surprise that we would expect from a model that is not knowledgeable about that specific language or domain.

\subsection{Addessing differences in tokenization}
\label{sec:tokenization}

The set of models $\mathcal{M}$ that we use allow us to compute per-token perplexity values, but these values are not directly comparable across models due to differences in tokenization.
For instance, while one model's tokenizer may transform ``Word'' into the single token `ĠWord', another model may tokenize it as `ĠWo' and `rd'.

We address this issue by grouping the tokens that correspond to the same word.
To do the aggregation, we choose to use the maximum perplexity value for each word's tokens.
We justify this choice on the causal nature of the language models we consider.
As each token's probability is conditioned on the tokens that precede it, the probability of the predicted first token of a word is typically lower --- and thus, the perplexity is higher --- than those of the subsequent tokens.

For instance, in the previous example, it is likely that the perplexity of the token `ĠWo' is higher than the perplexity of the token `rd'; it is easier to predict the second part of a word once the first part has been seen.
If we had chosen to use the average perplexity instead, models that tokenize words into more tokens would have shown lower perplexity values for the same words, which would have biased our results.
We therefore argue that the maximum perplexity value is a more robust choice for aggregating the perplexities of the tokens that correspond to the same word, and we use this approach in our method.

\subsection{Addressing differences in perplexity scales}
\label{sec:perplexity_scales}




After obtaining the word-level perplexities of the words in the answer, we face an additional challenge: perplexities are not generally comparable across models.
Specifically, the perplexity values can differ significantly across models due to differences in tokenization, model architectures and training data, which makes it difficult to compare the perplexities directly.

To address the differences in perplexity scales, we choose to normalize the perplexities based on the question perplexities so that they have a mean of 0 and a standard deviation of 1.
The calculation of the normalization parameters is done using exclusively the tokens that are part of the question, but the normalization is applied to all tokens.
This is because we desire the perplexities of the question tokens to be similar across models, given that we do not expect the question to be hallucinated.
At the same time, as the answer may contain multiple hallucinated spans, we expect that the normalization process will still preserve significant differences in perplexity values across words in different models, which we find to be the case in practice.

We also exclude the first $\text{min}(5, |t_{\text{q}}| - 2)$ of the $t_{\text{q}}$ question tokens from the calculation of the normalization parameters.
We do this to avoid the normalization being skewed by the higher perplexity values that are often found in the first few tokens of the question.

\subsection{Assigning weights to specialized models}
\label{sec:weights}

To assign weights to specialized models, we employ an instruction-following model, which we task with determining the most suitable model for dealing with the question-answer pair at hand.
We include the specific prompt used for this task in the supplementary material, which essentially asks the model to identify the specialization (model) that is most related to the question-answer pair.
We design the prompt so that the next token generated by the model corresponds to the name of the specialized model, a probability simplex over the model's vocabulary, which includes the probabilities for the tokens we assign to each specialized model.
We use these probabilities to assign weights to the specialized models by computing a softmax over the probabilities of the subset of tokens that correspond to the start of the word we associate with each model.

For instance, if as in Figure~\ref{fig:overview}, our specialized models are named ``Chinese'', ``English'', ``French'', ``German'', and ``Spanish'', and these names are associated with the tokens `ĠChinese', `ĠEnglish', `ĠFrench', `ĠGerman', and `ĠSpanish', we compute a softmax over the probabilities of these tokens to assign weights to the specialized models $\mathcal{M}$ so that we obtain $\alpha_j \forall j \in \{1, \ldots, |\mathcal{M}|\} = (0.01, 0.04, 0.01, 0.72, 0.22)$.

Additionally, we incorporate a temperature parameter $\tau$ into the softmax function: $\alpha_j = \frac{\exp(\text{logit}_j / \tau)}{\sum_{k}^{|\mathcal{M}|} \exp(\text{logit}_k / \tau)}$, where $\text{logit}_j$ is the logit associated with the first token of the name of ${m}_j$.
This controls the smoothness of the resulting distribution, which we optimize as part of the hyperparameter search process in Section~\ref{sec:hyperparameters}.

\subsection{Tagging hallucination spans}
\label{sec:tagging}

Our method allows to obtain a hallucination score $H$ for each word of a question-answer pair, but it does not directly tag spans of hallucinated text in the answer, such as the ``Philip II'' entity in Figure~\ref{fig:overview}.
Therefore, we propose a simple heuristic to tag the spans of hallucinated text from the $H$ scores.

\paragraph{Word selection.} We first select the words in the answer that have a $H$ score above $\sigma$ standard deviations from the mean $H$ score of the answer.
This threshold is a hyperparameter that we tune on the validation set, as we describe in Section~\ref{sec:hyperparameters}.

\paragraph{Span tagging.} We take the selected words and use our instruction-following model to predict the start and end of the hallucinated spans.
For each selected word, we feed the model with the full answer text and the selected word along with its direct preceding and following words.
We ask the model to use this as a reference to propose a span of hallucinated text, which should usually form a complete unit of meaning, such as a named entity or a clause.
We provide the full prompt in Appendix~\ref{sec:appendix}.
We use the response of the model to generate hard labels of the form $(s, e)$, where $s$ and $e$ are the start and end indices of the hallucinated span, respectively.

To generate soft labels, which are spans that are annotated with a probability of being hallucinated, we use the $H$ score of the word that was used as the reference for the model.
There is also a possibility that two hallucinated spans overlap, in which case we take the average of the probabilities that were assigned to each character --- for instance, if we have two overlapping spans $(s_1, e_1)$ and $(s_2, e_2)$ with probabilities $p_1$ and $p_2$, the probability of each character in the overlap is $\frac{p_1 + p_2}{2}$; while the probability of the characters outside the overlap is just the probability of the span that contains them.
This process generalizes to cases where more than two spans overlap.

\section{Experiments}
\label{sec:experiments}

We only utilize the original datasets provided by \citet{vazquez-etal-2025-mu-shroom}, which are divided into validation (25\%) and test (75\%) sets.

\subsection{Hyperparameter tuning}
\label{sec:hyperparameters}

We used the Optuna \cite{optuna} framework for hyperparameter tuning.
This allowed us to efficiently explore the hyperparameter space and select the best performing set of hyperparameters for our model.
Table \ref{tab:hyperparameters} shows the hyperparameters that we tuned and the ranges we explored, as well as the best values we found.
We exclusively used the validation set to tune the hyperparameters.
We utilized the English subset only, due to the computational cost of exploring all languages for hyperparameter tuning.

\begin{table}[h]
  \centering
  \begin{tabular}{lll}
    \toprule
    \textbf{Hyperparameter} & \textbf{Range} & \textbf{Best Value}     \\
    \midrule
    $\beta$                 & [0.0, 1.0]     & \num{0.4962175701}      \\
    $\sigma$                & [0.0, 2.0]     & \num{0.016472}          \\
    $\tau$                  & [0.0, 10.0]    & \num{3.392846885335018} \\
    \bottomrule
  \end{tabular}
  \caption{Hyperparameters tuned and their best values.}
  \label{tab:hyperparameters}
\end{table}

\subsection{Choice of models}
\label{sec:choice_of_models}

For our experiments, we used a set of specialized models that were trained on different languages, the Goldfish family by \citet{goldfish}.
We also used a model specialized in general knowledge, \texttt{facebook/xglm-7.5B} \cite{lin-etal-2022-shot}, for cases when a question is not specific to the cultural or linguistic domain of the other models, as well as a model specialized in mathematical reasoning, \texttt{DeepSeek-R1-Distill-Qwen-7B} \cite{deepseekai2025deepseekr1incentivizingreasoningcapability}.
These models, which are computationally intensive, can be omitted for efficiency --- our preliminary experiments did not show a significant degradation in performance when they were not used.
As an instruction-following model, we use \texttt{Llama-3.1-8B-Instruct} \cite{llama3modelcard}.

\section{Results}
\label{sec:results}

\begin{table}[h]
  \centering
  \begin{tabular}{lll}
    \toprule
    \textbf{Language} & \textbf{IoU}     & \textbf{Sp. Corr.} \\
    \midrule
    Arabic            & \num{0.30855946} & \num{0.29248316}   \\
    Basque            & \num{0.31071916} & \num{0.18597963}   \\
    Catalan           & \num{0.38380600} & \num{0.37765718}   \\
    Czech             & \num{0.25661336} & \num{0.24186781}   \\
    English           & \num{0.30361215} & \num{0.24108234}   \\
    Farsi             & \num{0.33040955} & \num{0.31872162}   \\
    Finnish           & \num{0.35908305} & \num{0.28587640}   \\
    French            & \num{0.31818551} & \num{0.27682063}   \\
    German            & \num{0.31450130} & \num{0.23420286}   \\
    Hindi             & \num{0.25726150} & \num{0.25536404}   \\
    Italian           & \num{0.41719988} & \num{0.42446976}   \\
    Mandarin          & \num{0.12603346} & \num{0.10749439}   \\
    Spanish           & \num{0.24021837} & \num{0.23512139}   \\
    Swedish           & \num{0.37804982} & \num{0.20815354}   \\
    \bottomrule
  \end{tabular}
  \caption{\textbf{Scores} for the 14 languages in the test set.}
  \label{tab:scores}
\end{table}

Table~\ref{tab:scores} shows the results of our method on the test sets.
We report two metrics, the Intersection-over-Union (IoU) and the Spearman correlation (Sp. Corr.) coefficient of our annotated spans with respect to the human annotations.
We make two main observations:

\paragraph{Performance.}
Our method achieves IoU scores of around \num{0.3} and comparable Spearman correlation values.
These scores are higher than a neural baseline (close to \num{0}), or marking all words as hallucinated (\num{0.3} and \num{0}, respectively), but when compared to other approaches in the literature --- which can achieve scores of around \num{0.6} for similar metrics \cite{vazquez-etal-2025-mu-shroom} --- there is still room for improvement, particularly by addressing the limitations we discuss in Section~\ref{sec:limitations}, a direction that we hope to explore in future work.

\paragraph{Multilinguality.}
We designed our approach with the goal of being able to work across multiple languages, which is reflected in the results.
In particular, the performance of our method does not degrade when considering languages other than English.
This is a promising result, as it suggests that our method is robust to the language of the input, and can be applied to a wide range of languages without significant loss of performance.

In general, we believe that our results show promise for a method that can detect hallucinations in language models across multiple languages by leveraging the divergence between the perplexities of specialized models.
This is an avenue that merits further exploration, by exploring alternative configurations and architectures, as well as by addressing the limitations we discuss in Section~\ref{sec:limitations}.

\section{Conclusion}
\label{sec:conclusion}

In this paper, we propose a method to detect hallucinations in language models by comparing the perplexities of a set of specialized models.
We leverage the divergence between the perplexities of the models to detect hallucinations, and we propose a heuristic to tag spans of hallucinated text in the answer.
Our method is designed to work across multiple languages, while there exists room for improvement, we show good generalization across languages, which is a promising result that can be built upon in future work.
Generally, we believe that our method can be a valuable tool for detecting hallucinations in language models, and we hope that it can contribute to the development of more reliable and trustworthy AI systems.

\newpage

\section*{Ethical considerations}
\label{sec:ethics}

The work presented in this paper is critical to ensure that models produce correct and valid answers, particularly in the context of tagging spans of hallucinations in multilingual settings. By focusing on detecting and tagging hallucinations in large language models, we aim to help develop a safer and more trustable AI. This is essential for the responsible deployment of AI systems in real-world applications, where the accuracy and reliability of the generated content are paramount.

Trustworthiness in language models is a significant concern, especially when these models are used across multiple languages and domains. The presence of hallucinations can lead to severe consequences, such as misinformation, loss of user trust, and potential harm. Therefore, it is imperative to develop methods that can effectively identify and tag these hallucinations to ensure that the outputs of language models are both accurate and reliable.

Our approach contributes to the broader goal of creating AI technologies that can benefit all users by providing more accurate and trustworthy outputs in multilingual contexts. By leveraging specialized model divergence, we can detect and tag hallucinated spans, thereby enhancing the overall user experience and fostering greater confidence in AI systems. This work not only advances the field of natural language processing but also supports the ethical deployment of AI, ensuring that these technologies serve as reliable tools for users across different languages and domains.

\section*{Limitations and future work}
\label{sec:limitations}

As with any research endeavor, our work is not without its limitations. We aim to contribute to the development of a responsible AI research community by openly discussing the potential shortcomings of our approach and outlining areas for future improvement:


\paragraph{Annotation inconsistency.}
The tagging of the examples exhibited considerable inconsistency, despite the annotators being provided with detailed instructions. To illustrate this point, consider the following example from the English validation set:

\begin{quote}
  \textbf{Question:} ``What did Petra van Staveren win a gold medal for?''

  \textbf{Answer:} ``Petra van Stoveren won a silver medal in the 2008 Summer Olympics in Beijing, China.''
\end{quote}

There could be multiple alternatives for tagging hallucinated spans in this example, depending on the annotator's interpretation of the text.
For instance, one might tag ``Stoveren'' as hallucinated, as the correct name is ``Staveren.''
Also, it'd be possible to tag ``silver'' as hallucinated, as the question explicitly mentions a gold medal.
Another evaluator might instead tag ``2008 Summer Olympics in Beijing, China'' as hallucinated, because Petra van Staveren won the gold medal at the 1984 Summer Olympics in Los Angeles \cite{petra_van_staveren}.
And yet another supervisor might indicate that the entire answer is hallucinated, as the question asks about \emph{what} she won a gold medal for (100 meter breaststroke), not \emph{where} she won it.

Therefore, while we acknowledge the subjective nature of the task, which makes it challenging to obtain a consistent ground truth, we also recognize that \textbf{our results are heavily influenced} by the structure of these human annotations and that this introduces \textbf{noise} into the scores and evaluation process.

\paragraph{Language of the inputs.}
Our method is designed to work across multiple languages, but each model's performance may vary depending on the language of the input. For example, if the input in Figure~\ref{fig:overview} --- which relates to a Spanish king --- was written in Hindi, and we processed it with the Spanish model, it may not be able to detect the hallucination (assuming that it was not trained in Hindi). This language specificity is something that we leverage as a strength in our approach, but can also prove to be a limitation in specific scenarios like this one.

This could be addressed by translating the input to each model's language, but this would introduce potential errors in the translation process, and would generate the challenge of associating the hallucinated spans back to the original text. Possible avenues to address this would be to align the words in different languages with their embedded representations, or to ask a multi-lingual model to identify what word in the original language corresponds to the hallucinated span in the translated text. But we expect this task to be far from trivial --- for instance, if the word ``king'' doesn't exist in a specific language, how could we know that the hallucination was in that word?

\paragraph{Languages $\neq$ countries.}
In this work, we have implicitly assumed that language-specialized models are tied to certain countries (\emph{e.g.}, French to France).
However, this is not always the case, as languages can be spoken in multiple countries, and countries can have multiple official languages.
We try to account for this by designing our weight assignment prompt to focus on the \emph{culture} related to a language, rather than a specific country, but further investigation is needed to understand if this significantly affects the performance of our method.


\newpage
\bibliography{custom}

\newpage
\appendix

\section{Supplementary materials}
\label{sec:appendix}

For the sake of brevity, we provide all supplementary materials at \url{https://github.com/ACMCMC/ask-a-local}.

\end{document}